\documentclass{article}

\usepackage{arxiv}
\usepackage{calligra}
\usepackage{amsmath,amssymb,amsfonts}
\usepackage[utf8]{inputenc} 
\usepackage[T1]{fontenc}    
\usepackage{hyperref}       
\usepackage{url}            
\usepackage{caption,booktabs}    
\usepackage{nicefrac}       
\usepackage{microtype}      
\usepackage{lipsum}
\usepackage{amssymb}
\usepackage{cite}
\usepackage{comment}

\usepackage{algorithmic}
\usepackage{graphicx}
\usepackage{textcomp}
\usepackage{xcolor}
\usepackage[normalem]{ulem}

\usepackage{pifont}
\def\BibTeX{{\rm B\kern-.05em{\sc i\kern-.025em b}\kern-.08em
    T\kern-.1667em\lower.7ex\hbox{E}\kern-.125emX}}
\usepackage{svg}
\usepackage{forest}
\usetikzlibrary{shadows}

\usepackage{calc}
\usepackage[final]{pdfpages}

\usepackage{titlesec}
\titleformat{\paragraph}
{\normalfont\normalsize\bfseries}{\theparagraph}{1em}{}
\titlespacing*{\paragraph}
{0pt}{3.25ex plus 1ex minus .2ex}{1.5ex plus .2ex}

\definecolor{xm_color}{RGB}{180, 180, 180}
\newcommand*\rot{\rotatebox{90}}

\newcommand{\cm}{\ding{51}}%
\newcommand{\xm}{\textcolor{xm_color}{\ding{55}}}%

\newcommand{\indep}{\rotatebox[origin=c]{90}{$\models$}}

\newif\ifCommentsAuthors

\CommentsAuthorsfalse
\ifCommentsAuthors
  
    \definecolor{myred}{rgb}{.8,.0,.0}
    
    \definecolor{myblue}{rgb}{0,0,.8}
    \newcommand{\commentmarc}[1]{\textcolor{myblue}{#1}}
    \definecolor{mcolor}{rgb}{0,0.5,0.1}
    \newcommand{\commentghys}[1]{\textcolor{mcolor}{#1}}
    \definecolor{mcolor}{rgb}{0.5,0.2,0.1}
    \newcommand{\commentjulz}[1]{\textcolor{mcolor}{#1}}
    \definecolor{mygreen}{rgb}{.0,.8,.0}
    \newcommand{\commentjo}[1]{\textcolor{mygreen}{#1}}
    \definecolor{mypurple}{rgb}{.6,.1,.6}
    \newcommand{\commentmeeting}[1]{\textcolor{mypurple}{#1}}

\else

    \definecolor{myred}{rgb}{.8,.0,.0}
    
    \newcommand{\commentmarc}[1]{}
    \newcommand{\commentghys}[1]{}
    \newcommand{\commentjulz}[1]{}
    \newcommand{\commentjo}[1]{}
    \newcommand{\commentmeeting}[1]{}
    
\fi

\usetikzlibrary{arrows,shapes,positioning,shadows,trees}

\tikzset{
  basic/.style  = {draw, text width=3cm, drop shadow, font=\scriptsize, rectangle},
  root/.style   = {basic, rounded corners=2pt, thin, align=center,
                   fill=black!30},
  level 2/.style = {basic, rounded corners=2pt, thin,align=center, fill=black!20,
                   text width=8em},
  level 3/.style = {basic, thin, align=left, fill=black!5, text width=7.1em}
}

\usepackage{tikz}
\usetikzlibrary{fit,positioning}

\usepackage{caption}
\usepackage{subcaption}

\begin{document}

\author{

  Marc-André Carbonneau\thanks{Equal contribution}\\
  Ubisoft - La Forge\\
  \texttt{marc-andre.carbonneau2@ubisoft.com} \\
  \And
  Julian Zaïdi\footnotemark[1]  \\
  Ubisoft - La Forge\\
  \texttt{julian.zaidi@ubisoft.com} \\
 
   \And
 Jonathan Boilard \\
  École de technologie supérieure\\
  \texttt{jboilard1994@gmail.com} \\
  \And
 Ghyslain Gagnon \\
  École de technologie supérieure\\
  \texttt{ghyslain.gagnon@etsmtl.ca} \\

}

\title{Measuring Disentanglement: A Review of Metrics}
\maketitle

\begin{abstract}
Learning to disentangle and represent factors of variation in data is an important problem in AI. While many advances have been made to learn these representations, it is still unclear how to quantify disentanglement. While several metrics exist, little is known on their implicit assumptions, what they truly measure, and their limits. In consequence, it is difficult to interpret results when comparing different representations. In this work, we survey supervised disentanglement metrics and thoroughly analyze them. We propose a new taxonomy in which all metrics fall into one of three families: intervention-based, predictor-based and information-based. We conduct extensive experiments in which we isolate properties of disentangled representations, allowing stratified comparison along several axes. From our experiment results and analysis, we provide insights on relations between disentangled representation properties. Finally, we share guidelines on how to measure disentanglement.
\end{abstract}


\section{Introduction}
In recent years, learning disentangled representations has attracted considerable attention from the machine learning community \cite{Higgins2016, Kim2018, Kim2019, Suter2019, Eastwood2018, Ridgeway2018, Kumar2018, Chen2018, Do2020, Li2020, Sepliarskaia2020, Duan2020, Liu2020, Ridgeway2016, Locatello2019Few, Locatello2019challenge, Locatello2019fairness, VanSteenkiste2019, Desjardins2012, chen2016infogan, Boulianne2020, Burgess2017, Mathieu2019, Thomas2017, Press2019}. A disentangled representation independently captures true underlying factors that explain the data. Such representations offer many advantages: when used on downstream tasks, they improve predictive performance \cite{Locatello2019challenge, Locatello2019Few}, reduce sample complexity \cite{Bengio2013, Scholkopf2012, VanSteenkiste2019, Ridgeway2018}, offer interpretability \cite{Bengio2013, Higgins2016}, improve fairness \cite{Locatello2019fairness} and have been identified as a way to overcome \textit{shortcut learning} \cite{Geirhos2020}.

Originally, disentanglement was evaluated by visual inspection, but recent research efforts have been devoted to propose metrics for more rigorous evaluations \cite{Higgins2016, Kim2018, Kim2019, Suter2019, Eastwood2018, Ridgeway2018, Kumar2018, Chen2018, Do2020, Li2020, Sepliarskaia2020, Duan2020, Liu2020}. Frequently, a new metric is proposed alongside a new representation learning method to highlight benefits not captured by existing metrics. Unfortunately, it is often unclear what these metrics quantify, and under which conditions they are appropriate \cite{Suter2019, Do2020, Locatello2019challenge, Abdi2019, Sepliarskaia2020}. Fair quantitative evaluation is important to assess research progress by comparing new representation learning methods with the state-of-the-art, but also equally important for practitioners when performing model selection and hyper-parameter tuning \cite{Duan2020}.

While most metrics correlate on simple data sets, they do not on more complex and realistic data\cite{Locatello2019challenge}. Moreover, this correlation does not mean that they lead to the selection of the same model, as observed in \cite{Suter2019, Locatello2019challenge, Abdi2019}. We highlight this problem in our experiments in Section \ref{exp_model_selection}. Having metrics that lead to different conclusions means that before choosing a model or a hyper-parameter setting, one must chose an appropriate metric for the application. This is not a trivial task because existing metrics measure different properties of disentanglement and make different, often implicit, assumptions of these properties. Moreover, these metrics are sometimes complex procedures themselves subject to hyper-parameter configuration. The goal of this paper is to provide some guidance to practitioners for selecting a metric given an application.

Very few papers discuss how to measure disentanglement. In \cite{Locatello2019challenge}, the authors conduct a large-scale study on disentanglement in the unsupervised setting. Their main conclusion is that disentangling predefined factors is impossible without inductive bias, and that random seeds and hyper-parameters have a greater impact on performance than the architecture of the studied models. They also conducted experiments to measure the degree of agreement of the six metrics used in the paper. They found that five of the six metrics correlate on the simple dSprites data set \cite{Higgins2016}, but only mildly on other more realistic data sets. Unfortunately, no interpretations is given onto why one of the metrics sometimes inversely correlates with the others, or why metrics measuring different properties strongly correlate. In \cite{Eastwood2018}, the authors propose a framework for the evaluation of disentangled representations. They identify three desirable properties of a disentangled representation: explicitness, compactness and modularity. They introduce the idea that these properties should be quantified separately, and propose a new metric decomposed in three parts. The key idea of measuring different properties separately is also advocated in \cite{Ridgeway2018}. The authors point out that one of the three properties, compactness, is of lesser interest in practical scenarios. We will discuss these properties in detail in Section \ref{representation_props}.

To our knowledge, \cite{Sepliarskaia2020} is the only study focusing on comparing metrics. The authors organize metrics based on the basic disentanglement properties they measure. Then, they verify that metrics assign a high score to all perfect representations and a low score to all representations that do not satisfy the measured property. Through demonstrations, they expose failure cases for several metrics. This constitutes a significant step towards the theoretical analysis of metrics in extreme cases. 

In this paper, we propose an in-depth analysis of supervised disentanglement metrics with real-world applications in mind. We establish a clear taxonomy of metric families and underline their strengths and shortcomings. We compare the metrics with respect to many practical considerations such as robustness to noise and hidden factors, nonlinear relationships, accuracy, calibration and computational efficiency. We conduct experiments that abstract the representation learning model and data, which allows us to generate representations for which we can accurately control and isolate the properties under study. Moreover, it also alleviates difficulties related to the identification of ground truth generative factors in data sets. We focus our analysis on \textit{supervised} metrics (i.e. metrics that require ground truth factors) since there exist very few  \textit{unsupervised} metrics\cite{Duan2020, Do2020, Liu2020}. To our knowledge, this is the first time that such extensive and fully controlled experiments are conducted, and that metrics are compared in depth.

\textbf{Contributions}:
    
    \begin{itemize}
        \item We carry out an extensive review of disentanglement metrics, where we expose underlying assumptions, implementation complexity and other practical considerations. 
        \item We establish a clear taxonomy of metric families and underline their strengths and short-comings.
        \item We conduct experiments that eliminate ambiguities introduced by learning algorithms and data sets to directly measure a metric's performance.
        \item We release our code to allow for the use of our metric implementations and the reproduction of our experiments\footnote{\url{https://github.com/ubisoft/ubisoft-laforge-DisentanglementMetrics}}.
        \item We provide recommendations for meaningful comparison between representations, as well as guidance for selecting appropriate metrics depending on the application context.
      
    \end{itemize}

The rest of the paper is organized as follows: We start by identifying desirable representation properties that we wish to quantify. In Section \ref{metric_property}, we define desirable characteristic for a metric. In Section \ref{section_metrics}, we survey existing metrics and present our taxonomy. In section \ref{section_experiments}, we present our experiments and their results. In Section \ref{section_discussion}, we discuss the implications of our findings and provide insight on how to measure disentanglement. We identify possible extensions to the paper in the conclusion.

\section{Properties of a Disentangled Representation}
\label{representation_props}
Before analyzing metrics, we discuss what constitutes a disentangled representation. While there is no unanimously accepted definition of disentanglement, most agree on two main aspects \cite{Schmidhuber1992, Bengio2013, Ridgeway2016, Bouchacourt2018, VanSteenkiste2019, Do2020}. First, the representation has to be distributed. This means that an input is a composition of explanatory factors and corresponds to a single point in the representation space. As in \cite{Eastwood2018} we call this point a \textbf{code} in the remainder of this paper. Each factor is encoded in separate dimensions of the code. In Section \ref{factor_ind}, we further discuss factor independence and its implications. Second, the representation should also encode relevant information for the downstream task. Depending on the application and the representation learning algorithm, the way codes and factors relate to each other may vary significantly. We discuss Information content in Section \ref{info_content}.


\subsection{Factor Independence in Representation} 
\label{factor_ind}

Factor independence means that variation in one factor does not affect other factors, i.e. there is no causal effect between them \cite{Peters2017}. In a disentangled representation factors are also independent in the representation space. In other words, a factor affects only a subset of the representation space, and only this factor affects this subspace. Most authors agree on the importance of this property which has different names (e.g., \textit{disentanglement} \cite{Eastwood2018}, \textit{modularity} \cite{Ridgeway2018}). In this paper, we use the naming convention of \cite{Ridgeway2018} and refer to this property as \textbf{modularity}.

Some authors argue that the subset of the representation space affected by a factor should be as small as possible. Ideally, only one dimension completely describes a factor. This property is called \textit{completeness} in \cite{Eastwood2018}, but is called \textit{compactness} in \cite{Ridgeway2018}. In this paper we refer to this property as \textbf{compactness}.

\commentmarc{groups of latent dimensions and generative factors provides more flexibility, for example when multiple latent dimen- sions are used to describe one phenomenon (Esmaeili et al., 2018) or \cite{Suter2019}}

The desirability of compactness relates to the type of factors for a given application. As argued in \cite{Ridgeway2018}, enforcing compactness may be counterproductive. A group of code dimensions provides more flexibility when describing complex factors \cite{Esmaeili2019}. For instance, if an angle is represented as a single value $\theta \in [0, 2\pi ]$, there is a discontinuity in the space at $2\pi$. Alternatively, if two dimensions encode the angle (e.g. $\sin{\theta}$ and $\cos{\theta}$), continuity is preserved. Furthermore, sometimes factors are complex concepts like facial expression \cite{Desjardins2012} or phonetic content \cite{Boulianne2020}, which are unlikely describable by a 1-dimensional space. As a second argument against enforcing strict compactness, Ridgeway and Mozer\cite{Ridgeway2018} explain that allowing redundancy in the learned latent space allows for different equivalent solutions, which facilitate model optimization from a practical standpoint.

\commentmarc{Choosing factor}
All metrics assume that a set of independent factors exists in the problem. However, in practice, identifying useful and interpretable independent factors represents a challenging task \cite{Mathieu2019}. Factors must be conceptually independent, but should also be statistically independent \cite{Kim2018}. This condition is hard to satisfy in real-world data sets where certain factor realizations tend to co-occur more than others \cite{Ridgeway2016, Trauble2020}. For example, in a data set of fruit images, we could be interested in two conceptually different factors: fruit type and color. However, color is statistically dependent on the fruit type. Dependent factors impact modularity score. If two factors were to share information, parts of the representation relate to both. The selection of factors is application-specific and is beyond the scope of this paper.

\subsection{Information Content}
\label{info_content}

\commentmarc{no information loss}
To be truly useful, a representation should completely describe explanatory factors of interest. In other words, it should be possible to retrieve the complete factor realization from a point in the representation space (code). We call this property \textbf{explicitness} as in \cite{Ridgeway2016}. 

Perfect explicitness entails that a generalizable relation between factors and codes was learned. The nature of this relation may vary and is implicitly assumed by some metrics. A linear relation between factors and codes is the simplest, and arguably most desirable, type of relation \cite{Ridgeway2018}. In applications such as user-specified conditioned generation, learning a monotonic relation, even if not linear, allows for intuitive navigation in the representation space. 

\commentmarc{modes in factors or categorical}
While a monotonic relation is a desirable characteristic for continuous factors, sometimes they are best described as categorical. In that case, the learning model must partition the representation space in regions corresponding to each category. The sample distribution in this space is multi-modal. Navigating this representation space by increasing the value of a code makes little sense. Measuring disentanglement with categorical factors necessitates metrics that do not make assumptions on the nature of the factor-code relation. 

\commentmarc{notes on invariance}
In \cite{Bengio2013}, the authors argue that a good representation should be invariant to other factors and noise. Unfortunately, it is not always clear which factors are pertinent for the downstream tasks. This is why it is advocated to learn as many factors as possible and discard as little information as possible \cite{Desjardins2012, Bengio2013}.

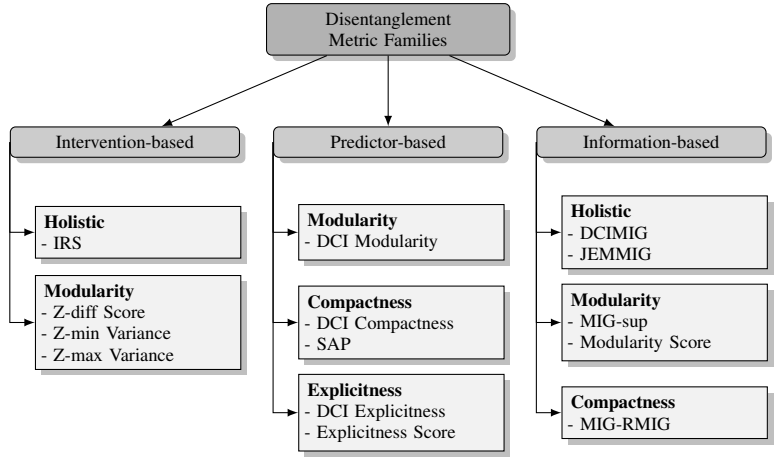
\begin{figure*}[ht]
\caption{Taxonomy of disentanglement metrics. Metrics are grouped in families based on their underlying working principle. Each family is divided in groups based on the disentanglement property that they are designed to measure.}
\centering

\begin{tikzpicture}[
  level 1/.style={sibling distance=35mm},
  edge from parent/.style={->,draw},
  >=latex]

\node[root] {Disentanglement Metric Families}
  child {node[level 2] (c1) {Intervention-based}}
  child {node[level 2] (c2) {Predictor-based}}
  child {node[level 2] (c3) {Information-based}};

\begin{scope}[every node/.style={level 3}, node distance=12mm]
\node [below of = c1, xshift=5pt] (c11) {\textbf{Holistic}\\ - IRS};
\node [below of = c11] (c12) {\textbf{Modularity}\\ - Z-diff Score \\  - Z-min Variance \\ - Z-max Variance};

\node [below of = c2, xshift=5pt] (c21) {\textbf{Modularity}\\ - DCI Modularity};
\node [below of = c21] (c22) {\textbf{Compactness}\\ - DCI Compactness \\ - SAP};
\node [below of = c22] (c23) {\textbf{Explicitness}\\ - DCI Explicitness\\ - Explicitness Score};

\node [below of = c3, xshift=5pt] (c31) {\textbf{Holistic}\\ - DCIMIG\\ - JEMMIG};
\node [below of = c31] (c32) {\textbf{Modularity}\\ - MIG-sup\\ - Modularity Score};
\node [below of = c32] (c33) {\textbf{Compactness}\\ - MIG-RMIG};
\end{scope}

\foreach \value in {1,...,2}
  \draw[->] (c1.west) |- (c1\value.west);

\foreach \value in {1,...,3}
  \draw[->] (c2.west) |- (c2\value.west);

\foreach \value in {1,...,3}
  \draw[->] (c3.west) |- (c3\value.west);
\end{tikzpicture}
\label{fig_taxonomy}
\end{figure*}

\section{Characterization of Disentanglement Metrics}
\label{metric_property}

Section \ref{representation_props} established three main properties of disentangled representations: modularity, compactness, and explicitness. This section enumerates desirable characteristics for a disentanglement metric. 

\commentmarc{Exact and calibrated} 
A metric should accurately measure a disentanglement property or a subset of the properties described in Section \ref{representation_props}. Ideally the metric should not have failure modes as identified in \cite{Sepliarskaia2020, Kim2018}. The scoring range of the metric should be calibrated. The metric should attribute the minimum score to a completely random or fully entangled representation and a perfect score to a perfectly disentangled representation. We verify this in the experiment of Section \ref{perfect_plus_noise}. We provide details on how to normalize the output range for metrics that were not normalized in their original implementation in Section \ref{section_metrics}.

\commentmarc{Gradually decreasing scores}
In addition to being calibrated, a metric score should also evolve linearly with the quality of the disentanglement properties that it measures. The worst-case scenario would be a metric that acts as a step function, which offers poor score interpretability and makes the comparison between two models with the same score meaningless. Moreover, such behavior renders the metric highly unstable. We evaluate how \textit{linearly} the metric scores change with respect to explicitness in Section \ref{perfect_plus_noise}, as well as to compactness and modularity in Section \ref{exp_decrease_mod_comp}. 

\commentmarc{Relations} As discussed in section \ref{info_content}, every metric makes an implicit assumption on the shape of the factor-code relationship. Sometimes, the application may dictate the type of relationship expected. For instance, if a linear relation is expected, metrics which penalize nonlinear relations \cite{Eastwood2018, Kumar2018} are best equipped to assess the quality of a representation. However, in most situations metrics should not make any assumptions and should capture nonlinear and multimodal relations. In Section \ref{exp_nonlinear} we compare metrics on monotonic and increasingly nonlinear relations.

\commentmarc{Sensitivity to hyper-parameters and number}
A metric should not be overly sensitive to hyper-parameter configuration. Low parameter sensitivity ensures stability across different configurations. A metric overly sensitive to configuration behaves unpredictably and may lead to inaccurate conclusions when comparing models. Examples if hyper-parameters include predictor parameters (e.g., regularization weights), discretization granularity, batch size, and validation protocol. The best way to mitigate hyper-parameter sensitivity is to reduce the number of parameters in the first place.    

\commentmarc{Noise Sensitivity} In real-world applications, data sets are likely to be noisy. Metrics that measure compactness or modularity should be tolerant to noise, while explicitness metrics should reflect the amount of noise in the representation. We measure robustness to noise in the experiment of Section \ref{perfect_plus_noise}. In the context of measuring disentanglement, explaining factors that are not targeted by the metric can also be viewed as sources of noise. In most real-world applications, there will be several factors that will not be identified and measured. We evaluate tolerance to these distracting factors in Section \ref{exp_partial_desc}.

Finally, there are practical considerations when choosing a metric. Some metrics have a high computational complexity, while others require a large number of data points to yield meaningful results. We briefly discuss these considerations in Section \ref{exp_sample_efficiency} and Section \ref{section_discussion_pratical}.

\section{Overview of Metrics}
\label{section_metrics}
This section surveys existing supervised metrics. We propose a new taxonomy that organizes metrics into three families. A family groups metrics based on their underlying working principle. \textbf{Intervention-based} metrics compare codes by creating subsets of data in which one or more factors are kept constant. \textbf{Predictor-based} metrics use regressors or classifiers to predict factors from codes. \textbf{Information-based} metrics leverage information theory principles, such as mutual information (MI), to quantify factor-code relationships. Inspired by \cite{Sepliarskaia2020}, we further divide each family in groups based on the disentanglement property that the metrics are designed to measure. Holistic methods capture two or more properties in a single score. Figure \ref{fig_taxonomy} shows all metrics organized following the proposed taxonomy. In the rest of this section, after introducing the notation, we go over all families in greater detail and describe metrics individually.

\begin{figure}[ht]
\caption{Illustration of the notation.}
\centering
\begin{tikzpicture}
\tikzstyle{main}=[circle, minimum size = 8mm, thick, draw =black!80, node distance = 16mm]
\tikzstyle{connect}=[-latex, thick]
\tikzstyle{box}=[rectangle, draw=black!100]
  \node[main, fill = white!100] (v) [label=center:\textbf{v}] { };
  \node[main] (x) [right=of v,label=center:\textbf{x}] { };
  \node[main] (z) [right=of x,label=center:\textbf{z}] {};
  \path (v) edge [connect] node[pos=0.5, above]{$g(.)$ } (x)
        (x)  edge [connect] node[pos=0.5, above]{$r(.)$ } (z);
		
\end{tikzpicture}
\label{notation_fig}
\end{figure}
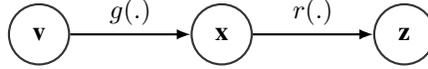

\subsection{Notation}

Inspired by \cite{Kim2019}, we denote a set of $N$ observations as $X = \{\textbf{\textup{x}}_1, \textbf{\textup{x}}_2, ..., \textbf{\textup{x}}_N \}$. Each observation is assumed to be completely explained by a set of $M$ factors $\mathcal{V} = \{v_1, v_2, ..., v_M \}$ through a generative process $g(\textbf{\textup{v}}) \mapsto \textbf{\textup{x}}$. We denote $V = \{\textbf{\textup{v}}_1, \textbf{\textup{v}}_2, ..., \textbf{\textup{v}}_N \}$ the set of factor realizations that produced $X$. A representation learning algorithm is a mapping $r(\textbf{\textup{x}})\mapsto \textbf{\textup{z}}$  where $ \textbf{\textup{z}} \in \mathbb{R}^{d}$ is a point in the learned code space denoted by $\mathcal{Z}= \{z_1, z_2, ..., z_d \}$. $Z = \{\textbf{\textup{z}}_1, \textbf{\textup{z}}_2, ..., \textbf{\textup{z}}_N \}$ is the set of all points in X projected in the code space by $r(.)$. Supervised disentanglement metrics compute a score by comparing $V$ to $Z$. Figure \ref{notation_fig} illustrates the notation. Throughout this paper, boldface lowercase letters represent vectors.


\subsection{Intervention-based Metrics}

The metrics in this family evaluate disentanglement by fixing factors and creating subsets of data points. Codes and factors in the subsets are compared to produce a score. To sample the fixed size data subsets, these methods discretize the factor space. This sampling procedure necessitates large quantities of diverse data samples to produce a meaningful score. The main advantage is that these metrics do not make any assumptions on the factor-code relations. However, there are several hyper-parameters to adjust such as the size and the number of data subsets, the discretization granularity, classifier hyper-parameters, or the choice of a distance function. Finally, \cite{Kim2018} and \cite{Sepliarskaia2020} identified several failure modes (i.e. situations where the metrics wrongly score representations).

\subsubsection{Z-diff}
\label{z_diff}

The Z-diff metric \cite{Higgins2016}, sometimes called the $\beta$-VAE metric, selects pairs of instances to create \textit{batches}. In a batch, a factor $v_i$ is chosen randomly. Then, a fixed number of pairs are formed with samples $\textup{\textbf{v}}^1$ and $\textup{\textbf{v}}^2$ that have the same value for the chosen factor ($v_i^1 = v_i^2$). Pairs are represented by the absolute difference of the codes associated with the samples ($\textup{\textbf{p}} = \left |  \textup{\textbf{z}}^1 - \textup{\textbf{z}}^2 \right |$). The intuition is that code dimensions associated with the fixed factor should have the same value, which means a smaller difference than the other code dimensions. The mean of all pair differences in the subset creates a point in a final training set. The process is repeated several times to constitute a sizable training set. Finally, a linear classifier is trained on the data set to predict which factor was fixed. The accuracy of the classifier is the Z-diff score. For a completely random classifier we expect an accuracy of $1/M$ where $M$ is the number of factors. This can be used to scale the output closer to the $[0, 1]$ range.




\subsubsection{Z-min Variance}
\label{z_min_var}

The Z-min Variance\footnote{We renamed the metric to avoid confusion with the model of the same name.} metric \cite{Kim2018}, also called the FactorVAE metric, was introduced to address some of the weaknesses of the Z-diff metric. The intuition is the same as for the Z-diff; code dimensions encoding a factor should be equal if the factor value is the same. First, all codes are normalized by their standard deviation computed over the complete data set. For a subset, a factor is randomly selected and fixed at a random value. The subset contains sampled instances for which the selected factor is fixed at the selected value. Variance is computed over the normalized codes in the subset. The code dimension with the lowest variance is associated to the fixed factor. Several subsets are created and the factor-code associations are used as data points in a majority vote classifier. The Z-min Variance score is the mean accuracy of the classifier. As for Z-diff, random classifier accuracy of $1/M$ can be used to scale the output closer to the $[0, 1]$ range.




\subsubsection{Z-max Variance}
\label{z_max_var}

Z-max Variance\footnotemark[2] metric \cite{Kim2019}, also known as R-FactorVAE, is similar to Z-min Variance. The main difference is the approach used to collect subsets of samples. Here all factor values are fixed except one. This time the intuition is that if all factors are the same except one, and code dimensions corresponding to the free factor should exhibit higher variance. A majority vote classifier is also used to compute the score, but it is the code dimension with the highest variance that is chosen as a training point.

\subsubsection{Interventional Robustness Score (IRS)}

IRS \cite{Suter2019} computes distances between sets of codes before and after an intervention on factor realizations. The intuition behind the metric is that changes in \textit{nuisance} factors should not impact code dimensions attributed to targeted factors. First a reference set is created from instances where realizations of target factors are fixed. Then a second set contains instances with the same targeted factor realization but different realizations of nuisance factors. The metric computes the distance (e.g. $\ell_2$) between the mean of code dimensions associated to targeted factors. This sampling and distance measurement procedure is repeated several times and the maximum observed distance is reported. The final metric reports a weighted average of the maximum distances. The distances are weighted by the frequency of the factor realizations in the data set.





\subsection{Predictor-based Metrics}
These metrics train regressors or classifiers to predict factor realizations from codes ($f(\textbf{\textup{z}})\mapsto \textbf{\textup{v}}$). Then the predictor is analyzed to assess the usefulness of each code dimension in predicting the factors. These methods are naturally suited to measure explicitness. They are typically equipped to deal with continuous factors as well as categorical factors simply by choosing an appropriate predictor. However, compared to Information-based metrics, they require more design choices and hyper-parameter tuning. This means a metric is more likely to behave differently from one implementation to another.

\subsubsection{Disentanglement, Completeness and Informativeness (DCI)}
In \cite{Eastwood2018}, the authors propose a complete framework to evaluate disentangled representations instead of a single metric. They report separate scores for modularity, compactness and explicitness, which they call disentanglement, completeness and informativeness. Regressors are trained to predict factors from codes. Modularity and compactness are estimated by inspecting the regressor's inner parameters to infer predictive importance weights $R_{ij}$ for each factor and code dimension pair. They use a linear lasso regressor or a random forest for nonlinear factor-code mappings. For the lasso regressor, the importance weights $R_{ij}$ are the magnitudes of the weights learned by the model, while the Gini importance \cite{breiman2001} of code dimensions is used with random forests. 

The compactness for factor $v_i$ is given by $C_i = 1 + \sum_{j=1}^{d} p_{ij}\textup{log}_d\,p_{ij}$ where $p_{ij}$ is the \textit{probability} that code dimension $z_j$ is important to predict $v_i$. These probabilities are obtained by dividing each importance weight by the sum of all importance weights related to this factor: $p_{ij} = R_{ij} / \sum_{k=1}^{d}R_{ik}$. The compactness of the whole representation is the average compactness over all factors.

Similarly, the modularity for code dimension $z_j$ is given by $D_j = 1 + \sum_{i=1}^{M} p_{ij}\textup{log}_M\,p_{ij}$ where $p_{ij}$ the is \textit{probability} that code dimension $z_j$ is important to predict only $v_i$. This time the importance weights are normalized with respect to codes: $p_{ij} = R_{ij} / \sum_{k=1}^{M}R_{kj}$. The modularity score for the whole representation is a weighted average of the individual code dimension modularity scores $\sum_{j=1}^{d} \rho_j D_j$. The scores are weighted by $\rho_j$ to account for codes that are less important to predict factors. The weight $\rho_j$ is the total importance for $z_j$ normalized by the sum of all importance weights: $\rho_j = \sum_{i=1}^{M}R_{ij}/ \sum_{k=1}^{d}\sum_{i=1}^{M}R_{ik}$.

The prediction error of the regressor measures the explicitness of the representation. With normalized inputs and outputs, it is possible to compute the estimation error for a completely random mapping and use it to normalize the score between 0 and 1. We postulate that a representation is not explicit if the mean squared error (MSE) of the predictor is higher than the expected MSE between two uniformally distributed random variables ($X$ and $Y$). It can be showed that $\textup{MSE} = \mathbb{E}[(X-Y)^2] = 1/6$. Thus, explicitness can be written as $1- 6 \cdot \textup{MSE}$. In our implementation values under 0 are reported as 0.





\subsubsection{Explicitness Score}
In \cite{Ridgeway2018}, the authors propose to use a classifier trained on the entire latent code to predict factor classes, assuming that factors have discrete values. They suggest using a simple classifier such as logistic regression and report classification performance using the area under the ROC curve (AUC-ROC). The final score is the average AUC-ROC over all classes for all factors. The AUC-ROC minimal value is 0.5 which means that the score needs to be normalized to obtain a value between 0 and 1. In our implementation we balance weights in the loss of the logistic regression to account for class imbalance.





\subsubsection{Attribute Predictability Score (SAP)}
SAP \cite{Kumar2018} attributes a score $S_{ij}$ to all pairs of factor $v_i$ and code dimension $z_j$. A linear regression predicts a continuous factor from each code and $S_{ij}$ is the $R^2$ score of the regression. For categorical factors, it fits a decision tree on codes and reports balanced classification accuracy. Scores corresponding to codes with energy below a user specified threshold (i.e. \textit{dead-codes}) are set to 0. The final SAP score is obtained by computing the difference between the two highest $S_{ij}$ for all factors:
\begin{equation}
\textup{SAP} = \frac{1}{M} \sum_{i}^{M} S_{i\star} - S_{i\circ} 
\end{equation}
In this equation, $S_{i\star}$ is the highest score for factor $v_i$, while $S_{i\circ}$ is the second highest. $M$ is the number of factors. Similar $S_{i\star}$ and $S_{i\circ}$ means that explicitness is low if both values are low. Two similarly high values indicate that more than one code dimension encodes the factor which means low compactness. This corresponds to the \textit{gap} idea in MIG (Section \ref{sect_mig}).



\subsection{Information-based Metrics}
Information-based metrics compute a disentanglement score by estimating the mutual information (MI) between the factors and the codes. These methods require fewer hyper-parameters than intervention-based and predictor-based metrics. Moreover, they do not make assumptions on the nature of the factor-code relations. 

While elegant in theory the estimation of entropy and MI is non-trivial in practice. Even assessing the quality of the estimators remains an open problem \cite{Paninski2003}. Aside from specific cases where the distribution of the spaces is known and simple, it requires quantization of both spaces or a sampling procedure which needs to be parameterized. Most existing public MI-based metric implementations use the maximum likelihood estimator. For example in the widely used disentanglement\_lib\footnote{https://github.com/google-research/disentanglement\_lib} MI is computed as follows:
\begin{equation}
    I(v, z) = \sum_{i=1}^{B_v}\sum_{j=1}^{B_z}P(i,j)\log\left(\frac{P(i,j)}{P(i)P(j)}\right)
\end{equation}
Factor and code spaces are discretized in $B_v$ and $B_z$ bins. $P(i)$ and $P(j)$ are estimated as the proportion of samples assigned to bin $i$ and $j$ respectively over all samples ($N$). Similarly $P(i,j)$ is the proportion of samples assigned to both bin $i$ and $j$. 
Problems arise when estimating from under-sampled high dimensional data \cite{Hausser2009}. This is the case when computing MI, or joint entropy, between a factor and more than one dimension of the latent space. Moreover, the estimated MI value is affected by the granularity of the discretization which makes the metrics sensitive to this parameter. 

\subsubsection{Mutual Information Gap (MIG)}
\label{sect_mig}

MIG \cite{Chen2018} computes the MI between each code and factor $I(v_i, z_j)$. Then the code dimension with maximum MI is identified $I(v_i, z_\star)$ for each factor. Next, the second highest MI, $I(v_i, z_\circ)$, is subtracted from this maximal value. This difference constitutes the \textit{gap}. The gap is then normalized by the entropy of the factor:
\begin{equation}
\textup{MIG} = 
\frac{I( v_i, z_\star) - I( v_i, z_\circ)}{H(v_i)}
\end{equation}
The MIG score of all factors are averaged to report one score.

Robust MIG (RMIG) was proposed in \cite{Do2020}. It is identical to MIG in essence, but proposes a more robust formulation when MI is computed from the input space, which does not apply in our context. For the remainder of the paper we will refer to both MIG and RMIG as MIG-RMIG because our results apply to both in the same way.






\subsubsection{Joint Entropy Minus Mutual Information Gap (JEMMIG)}
MIG verifies that the information related to a factor is expressed by only one code dimension (compactness). However, modularity is not directly measured. For instance a code dimension could contain information about more than one factor. JEMMIG \cite{Do2020} addresses this drawback by including the joint entropy of the factor and its best code. 
\begin{equation}
\textup{JEMMIG} = H(v_i, z_\star) - I(v_i, z_\star) + I( v_i, z_\circ)
\end{equation}

As opposed to MIG, this metric indicates a high disentanglement quality with a lower score. The maximum value is bounded by $H(v_i) + \text{log}(B_z)$, where $B_z$ is the number of bins used in the code space discretization. This means that JEMMIG can be rewritten as follows to get a score between 0 and 1:
\begin{equation}
\widehat{\textup{JEMMIG}} = 1-\frac{H(v_i, z_\star) - I(v_i, z_\star) + I( v_i, z_\circ)}{H(v_i) + \text{log}(B_z)}
\end{equation}
As done for MIG, JEMMIG is reported as the average for all factors $v_i$.





\subsubsection{MIG-sup}
\label{sect_migsup}
MIG-sup \cite{Li2020} is an extension of MIG. As JEMMIG, it addresses the fact that MIG measures compactness, but does not measure modularity. It is designed to be used in conjunction with MIG. The idea is similar to MIG except that the MI gap is computed from the code point-of-view:
\begin{equation}
\textup{MIG-sup} = I(z_j, v_\star)-I(z_j, v_\circ)
\end{equation}
where $v_\star$ is the factor that has the highest MI with code dimension $z_j$. $v_\circ$ is the factor that has the second highest MI with code dimension $z_j$. $I(z_j, v_i)$ is the MI normalized by the entropy of the factor $v_i$. MIG-sup is reported as the average gap over all meaningful code dimensions. Meaningful dimensions can be identified by comparing the magnitude of $I(z_j, v_\star)$ for every code dimensions. In real-world scenarios a threshold has to be set to decide which code dimension is meaningful. The same representation would obtain largely different scores depending how selective is the threshold. Unfortunately there are no objective way of setting this threshold unless we know the correct disentanglement score in advance. The authors of MIG-SUP did not provide detail on the meaningful code dimension selection scheme used in the paper. To avoid thresholding, the code dimension selection process could be replaced with a scaling method inspired from DCI \cite{Eastwood2018}. In our implementation we consider all code dimensions.

\subsubsection{Modularity Score}
To measure modularity, in \cite{Ridgeway2018} the factor $v_\star$ which shares the maximum MI for each code dimension $z_j$ is identified. This maximal MI value $I(v_\star, z_j)$ is then compared with MI values of all other factors:
\begin{equation}
    \textup{modularity} = 1-\frac{\sum_{i \in \mathcal{V}_{\neq\star}}I(i,z_j)^2}{I(v_\star,z_j)^2(M-1)}
\end{equation}
We denote $\mathcal{V}_{\neq\star}$ as the set of all factors except $v_{\star}$ and $M$ as the number of factors. The average modularity score over all codes is reported.

\subsubsection{DCIMIG}
DCIMIG \cite{Sepliarskaia2020} is a metric inspired by DCI and MIG. As MIG, it computes MI gaps between factors and code dimensions. As DCI it analyzes a factor-code importance matrix. However, unlike DCI, DCIMIG reports a single score for all three disentanglement properties. DCIMIG starts by computing the MI between each factor and code dimension $I(v_i, z_j)$. Then the factor with maximum MI, $I(v_\star, z_j)$, is identified for each code. After, the second highest MI, $I(v_\circ, z_j)$, is subtracted from this maximal value. Thus we obtain a \textit{gap} for each code dimension $R_j = I(v_{\star}, z_j) - I(v_\circ, z_j)$. Each of these gaps $R_j$ relates to a code dimension and the factor for which MI is maximal. For each factor $v_i$, we find all associated gaps $R_j$ and use them as score $S_i$ for this factor. If there are more than one $R_j$ associated with the factor, $S_i$ equals the highest $R_j$. If there are none, $S_i = 0$. Finally the metric is the sum of all scores normalized by the total factor entropy:
\begin{equation}
    \textup{DCIMIG} = \frac{\sum_{i=1}^{M}S_i}{\sum_{i=1}^{M}H(v_i)}
\end{equation}

\section{Experiments}
\label{section_experiments}
In the experiments, we abstract the relation $\textbf{\textup{z}} = r(g(\textbf{\textup{v}}))$ by $\textbf{\textup{z}} = f(\textbf{\textup{v}})$. Except for Section \ref{exp_model_selection}, we do not learn representations on data sets, but instead we directly define $f(\textbf{\textup{v}})$ as a function that allows for complete control over the parameters of the representation evaluated by the metrics. This removes any ambiguities related to the quality of the data set, the choice of learning algorithm and its training, as well as the choice of factors to disentangle. We assume factors are selected following $v_i \indep v_j, i \neq j $ as prescribed in \cite{Suter2019}. The code for the experiments is publicly available\footnotemark[1].

\subsection{Model Selection}
\label{exp_model_selection}
\begin{figure}[ht]
     \centering
     \begin{subfigure}[b]{0.48\textwidth}
         \centering
         \includegraphics[width=\textwidth]{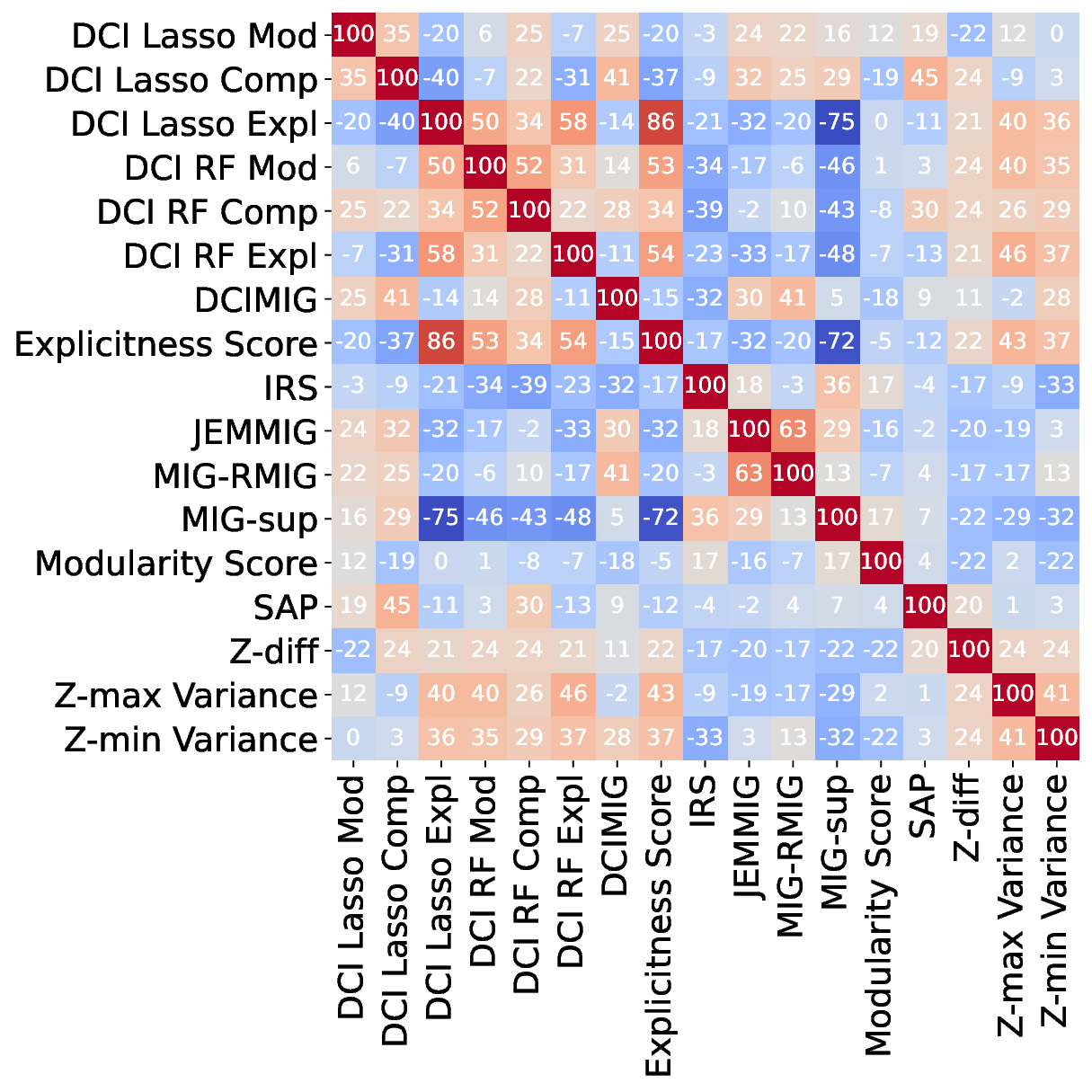}
         \caption{Cars3D}
     \end{subfigure}
     \begin{subfigure}[b]{0.48\textwidth}
         \centering
         \includegraphics[width=\textwidth]{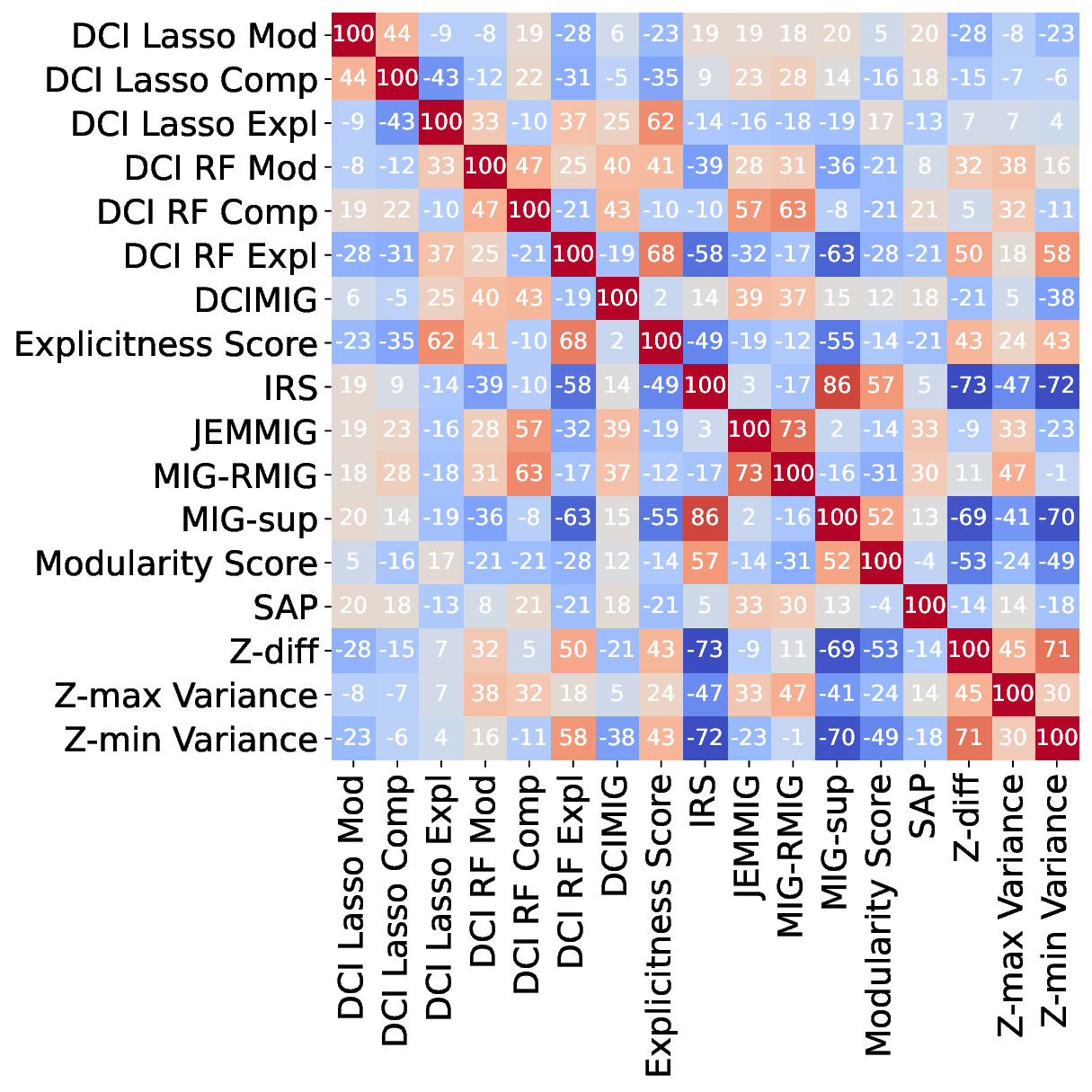}
         \caption{SmallNORB}
     \end{subfigure}
\caption{Kendall rank correlation coefficient ($\times 100$) between metrics for model rankings.}
\label{fig_kendall}
\end{figure}

In this experiment we validate that using different metrics to perform model selection or hyper-parameter tuning leads to different outcomes. The experiment emulates practitioners trying to tune model hyper-parameters to maximize disentanglement. We perform a grid search using the metric scores as the maximization objective. We arbitrarily chose to optimize two hyper-parameters for $\beta$-VAE \cite{Higgins2016}: The regularization strength ($\beta \in \{0.001, 0.01, 0.1, 1, 10, 100\}$) and the dimensionality of the representation space ($d \in \{2, 4, 8, 16, 32, 64\}$), which leads to 36 different hyper-parameter configurations. We use two standard data sets Cars3D \cite{Reed2015} and SmallNORB \cite{LeCun2004}. For each hyper-parameter configuration, we train the model for 300k training steps using the Adam optimizer and a batch size of 64 similar to \cite{Locatello2019challenge}. After training, we obtain 36 learned representations, one for each hyper-parameter configuration. We produce a ranking of the learned representations based on the scores obtained with each metric. Then, we measure the agreement between rankings for each pair of metrics with the Kendall rank correlation coefficient \cite{Kendall}. We report results in Figure \ref{fig_kendall}.

Our results show that two practitioners would have chosen different models if they measured disentanglement with different metrics, even if they were designed to quantify the same properties. This is consistent with results obtained in \cite{Suter2019, Locatello2019challenge, Abdi2019}. In Figure \ref{fig_kendall} we observe that for the same data set some metrics correlate, but often correlation is weak or even inverse. When comparing correlations across the two data sets, we observe a general correspondence in correlation directions, especially for strong correlations. This is reassuring since it indicates that the metrics are quantifying the same properties somewhat consistently across data sets. However, the magnitudes of these correlations vary indicating that there is an interplay between the nature of the data, the learned representations and the metric behaviours.

The experiment objective is two-fold. First, it confirms that the metrics are not equivalent and measure different properties under different assumptions, which motivates the present study. Second, it shows that it would be hazardous to compare metrics on representation learned from data because the exact factor-code relations are unknown. This is why we abstract the data and the learning process by devising fully parameterized relations for the subsequent experiments.

\subsection{Perfect Disentangled Representation with Noise}
\label{perfect_plus_noise}
\begin{figure*}[!ht]
\centering
\includegraphics[width=\textwidth]{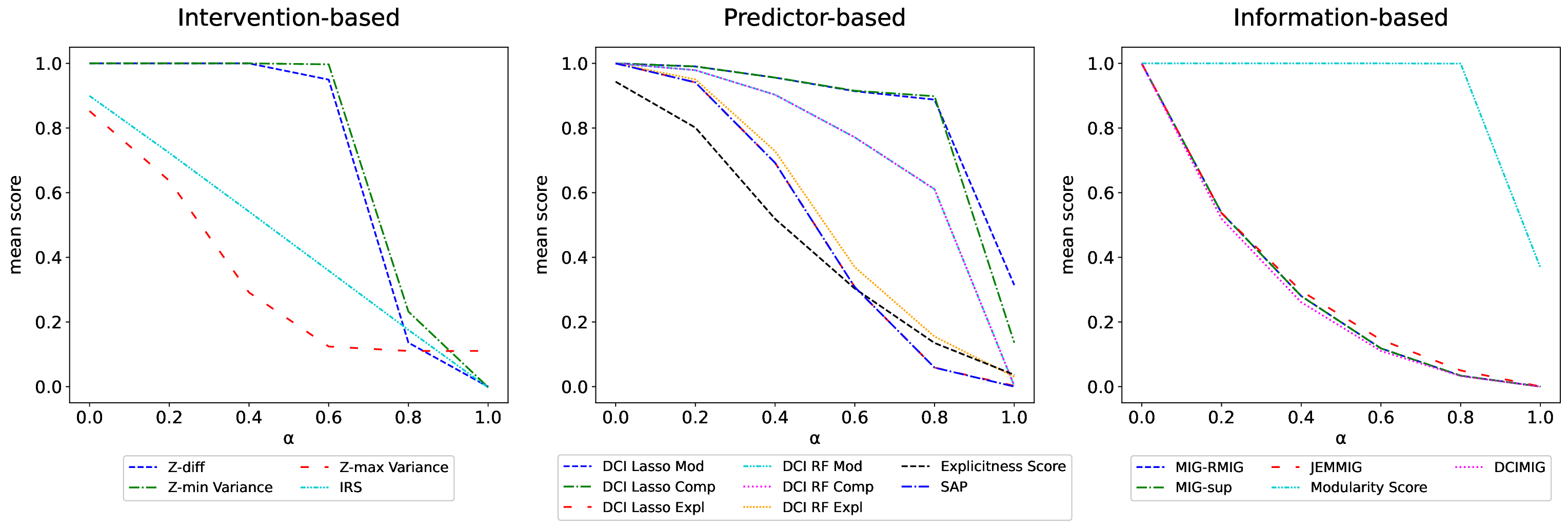}
\caption{Metric scores for perfectly disentangled representations under increasing noise level ($\alpha$).}
\label{fig_all_exp1}
\end{figure*}

In this section we evaluate how metrics behave in a scenario where we gradually depart from a perfect disentangled representation to a completely random representation. In a perfect representation, factors completely describe the data and have a one-to-one relation with codes. This scenario shows how metrics behave as explicitness decreases under perfect compactness and modularity. We also verify metrics are well calibrated (i.e. attribute a perfect score to perfect representation and a low score to noise). The factor-code relation is defined by:
\begin{equation}
     \textbf{\textup{z}} = f(\textbf{\textup{v}}) = (1-\alpha) \textbf{\textup{v}} + \alpha \textbf{\textup{n}}
\end{equation}
where $\textbf{\textup{n}} \sim \mathcal{U}(0,1)$, $\alpha \in [0, 1]$ and $ \textbf{\textup{v}}, \textbf{\textup{z}} \in \mathbb{R}^{M=d}$. We simulate a problem with 8 factors ($d = M = 8$). We tried different number of factors and found conclusions to be similar. Given a set of factor realizations $V$ we use $f(.)$ to obtain its representation in the code space $Z$. The set $V$ contains $N = 20k$ samples from the uniform distribution. We use the same 20k samples for all metrics. When necessary, factor values are discretized into 10 equal bins. We evaluate $\alpha$ at $\{0.0, 0.2, 0.4, ..., 1.0\}$. We repeat the experiment with 100 different sampled versions of $V$ using 100 random seeds and report the average result. Figure \ref{fig_all_exp1} shows the mean score for all metrics as the noise level ($\alpha$) increases. 

\commentmarc{Calibration}
Most metrics recognize a perfect representation and attribute a perfect score. There are three exceptions. IRS is unlikely to produce a perfect score for any representation because it computes a distance between codes for factors that are \textit{binned} together. Factors in the same bin are likely to differ within the range of the bin, which in turn results in small distances in code values for the same factor bin. This explains why IRS cannot attribute a perfect score to a perfect representation. To circumvent this problem, smaller discretization needs to be applied if the number of samples is large enough for the given application. The Explicitness score is the average of AUC-ROC for $M \times 10 = 80$ logistic regression classifiers trained in a one-versus-the-rest strategy. One classifier is trained for each bin value per factor. The optimizer does not consistently find the optimal solution for all classifiers which leads to an AUC-ROC under 1. Z-max Variance requires a dense combination of factor values to sample meaningful batches for the majority vote classifier. The 20k examples used in the experiment, when discretized in 10 bins, do not provide enough examples for a same factors realization. The direct consequence is a biased estimation of the variance, which causes a score under 1 for a perfect representation and a score higher than 0 for a completely random representation. To circumvent this problem, coarser discretization needs to be applied, which in turn might lead to an overestimation of the scores.

The majority of metrics attribute a score near 0 to complete noise. However, DCI for modularity and compactness scores the representation over 0.3 when using a lasso regressor. Even if the regressor accuracy is low, weights are still learned and compared to compute compactness and modularity scores. The regularization term in lasso pushes some weights towards 0 and thus sizable differences between them will be observed. This leads to observing random isolated factor-code relations which drive the score up. When using the Modularity score, the MI between each factor and code dimension should be similar. However, maximal MI value normalizes the score, which leads to a wrongfully optimistic value in most experiments in this paper.

\commentmarc{accurate explicitness}
When measuring explicitness under noise, an ideal metric score should steadily decrease as the noise level increases. IRS is a perfect example of a score that decreases linearly with noise. In fact, most metrics that focus on explicitness perform adequately. If explicitness metric scores should decrease in the presence of noise, we expect a different behavior from modularity or compactness metrics. Ideally, a metric should recognize these disentanglement properties, even in noisy representations. The predictor-based DCI exhibits a high noise robustness, which makes sense since predictors naturally discard noise information to improve generalization. This being said, their tendency to observe random isolated factor-code relations discussed above inflates this perception of noise robustness. In addition to being well calibrated, intervention-based metrics Z-diff and Z-min Variance also proved to be quite robust. Inversely, this experiment exposes the vulnerability to noise of information-based metrics. Noise causes codes to be assigned to neighbouring bins which decreases the observed MI between factors and codes.

\subsection{Decreasing Compactness and Modularity}
\label{exp_decrease_mod_comp}
In the previous section we observed how metrics behave as explicitness decreases. Now, we study what happens as we gradually decrease compactness and modularity, while explicitness remains perfect. The embedding function is constructed given by $\textbf{\textup{z}} = f(\textbf{\textup{v}}) = \textbf{\textup{v}}R$. The projection matrix $R$ is defined by:
\begin{equation*}
R = 
\begin{bmatrix}
1-\alpha & \alpha & 0 & \cdots & 0 \\
0 & 1-\alpha & \alpha & \cdots & 0 \\
0 & 0 & 1-\alpha &\cdots & 0\\
\vdots  & \vdots & \vdots  & \ddots & \vdots  \\
\alpha & 0 & 0 & \cdots & 1-\alpha 
\end{bmatrix}
\end{equation*}

\begin{figure*}[ht]
\centering
\includegraphics[width=\textwidth]{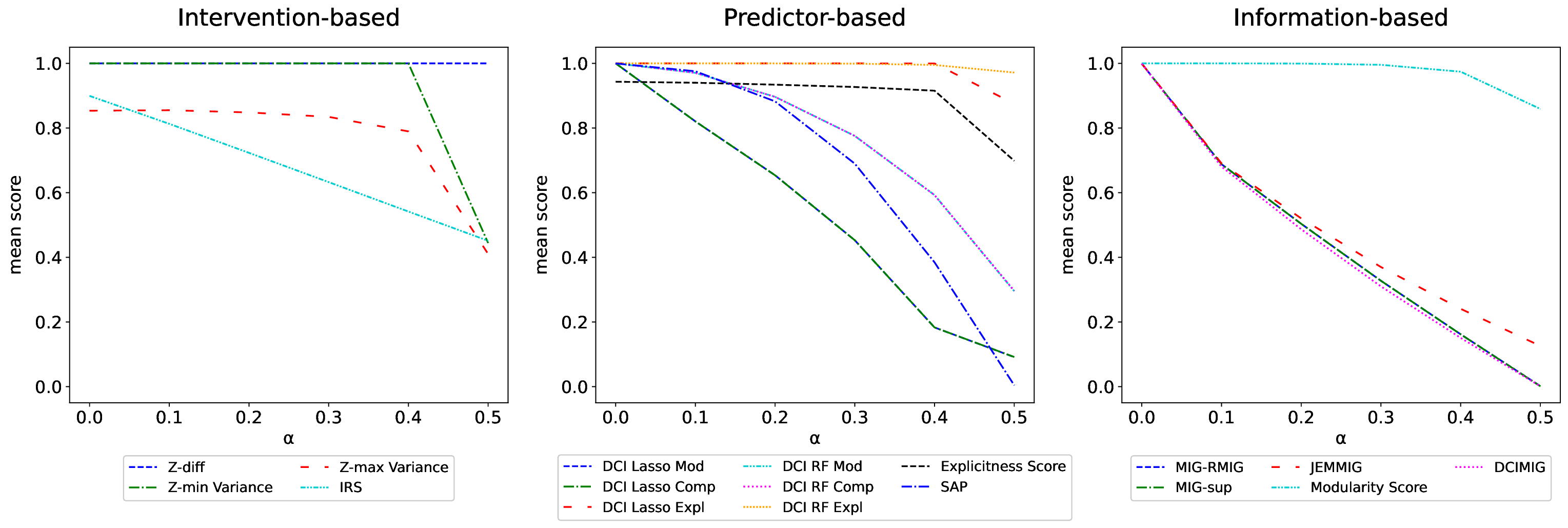}
\caption{Mean metric scores as the representation becomes less modular and less compact.}
\label{fig_space_rotation}
\end{figure*}

When $\alpha=0$, $R$ is the identity matrix and the representation is perfectly compact and modular. As $\alpha$ increases, all factors are represented by two code dimensions and each code dimension relates to two factors. Figure \ref{fig_space_rotation} shows how metric scores evolve as the representation becomes less modular and less compact.

Results from this experiment reveal several differences amongst metrics. Since the representation allows for complete recovery factor values, explicitness metrics maintain a high score as expected. We would expect modularity and compactness metric scores to linearly decrease as $\alpha$ increases. This is the case for most information-based metrics and predictor-based metrics. Interestingly, some metrics output a 0 score when a code dimension relates to two factors or \textit{vice versa}. This means that these metrics make no distinction between a representation where a code dimension relates to two factors and a representation where a code dimension relates to all factors. DCI is the best equipped metric to quantify this distinction because it will never yield a zero score unless all codes are equally relevant to predict the factors. The experiment also reveals a failure mode of intervention-based metrics as identified in \cite{Sepliarskaia2020}. These metrics consistently attribute a high score to the representations even when it is imperfect. The worst case is Z-diff that attributes a perfect score even when $\alpha=0.5$. It is always trivial for the classifier to identify a factor by finding the distinct combination of two code dimensions with the lowest difference.

\subsection{Modular but not Compact}
\label{exp_modular_not_compact}
Here we evaluate how the metrics behave when the representation is perfectly explicit and modular but not compact. As discussed in Section \ref{factor_ind}, compactness is of lesser interest than modularity in many real-world applications. Thus, it is important to assess the ability of the metrics to recognize modularity even when several code dimensions are used to describe a single factor. 

 The first experiment of this section emulates a model that has learned a decomposed representation of angles. When a scalar defines an angle, the representation space has a discontinuity at $2\pi$. Decomposing angles in sine and cosine values ensures the space is continuous which is preferred in many applications. Here, each factor represents an angle $\theta \in [0, 2\pi[$. Codes represent angles as $\text{cos} \, \theta$ and $\text{sin} \, \theta$. Factor realizations define four angles: $\textbf{\textup{v}} = [\theta_1, \theta_2, \theta_3, \theta_4]$ and the corresponding codes are given by $\textbf{\textup{z}} = [\text{cos}\,\theta_1, \text{sin}\,\theta_1,\text{cos}\,\theta_2, ..., \text{sin}\,\theta_4]$. Factor values are discretized to 10 bins ($v_i \in \{0, \pi/5, 2\pi/5, ..., 9\pi/5\}$). 

Following the same idea, we create a second data set where factors are encoded by two code dimensions. However, this time, factor-code relations are linear. This corresponds to a scenario where the representation learning algorithm has learned redundant codes. This scenario allows for comparison of results obtained in the previous experiment without having to account for the nonlinear relations (sine and cosine). We keep the same four factors, but we use linear relations:  $\textbf{\textup{v}} = [\theta_1, \theta_2, \theta_3, \theta_4]$ corresponds to $\textbf{\textup{z}} = [\theta_1, \theta_1,\theta_2, ..., \theta_4]$. 

Finally, we repeat the same experiment except that there are only two factors associated with four code dimensions each: $\textbf{\textup{v}} = [\theta_1, \theta_2]$ corresponds to $\textbf{\textup{z}} = [\theta_1, \theta_1,\theta_1, ..., \theta_2]$. Following the sampling methodology described in Section \ref{perfect_plus_noise}, we compute the metrics and report scores in Table \ref{table_angle}.

On the left of the result table, we can see that none of the intervention-based metrics penalizes representations for not being compact. This is in accordance with the results from the previous experiment. Predictor-based metrics exhibit different behaviors depending on the type of predictor used. The lasso predictor, unsurprisingly, has trouble dealing with the nonlinear sine and cosine relations. More interestingly, it has problems dealing with redundant codes. Since only one code dimension is necessary to predict a factor, the information from the duplicated code dimensions is discarded, encouraged by the regularization term. This falsely leads the metric to think that only one code dimension is associated with the factor, hence the perfect compactness for all experiments. Using a random forest predictor overcomes this problem. As observed in the preceding experiment, SAP and MIG which measure compactness cannot express to what degree a representation is not compact. This is because they compute a \textit{gap} which subtracts the two most significant terms and ignores all of the others. The same can be said for JEMMIG. DCIMIG while intended as a holistic method does not penalize non-compactness in this experiment. Finally, we can observe that information-based metrics have trouble dealing with nonlinear relations. This will be discussed in greater detail in the next experiment.

\begin{table*} \centering
\caption{Scores attributed to disentanglement where a factor is encoded with more than one code.}
\scalebox{0.92}{
\begin{tabular}{l|cccc|cccccccc|ccccc|}

&\rot{Z-diff} & 
\rot{Z-min Variance} & 
\rot{Z-max Variance} & 
\rot{IRS} &

\rot{DCI Lasso Modularity} &
\rot{DCI Lasso Compactness} &
\rot{DCI Lasso Explicitness} &
\rot{DCI RF Modularity} &
\rot{DCI RF Compactness} &
\rot{DCI RF Explicitness} &
\rot{Explicitness Score} &
\rot{SAP} &

\rot{MIG-RMIG} & 
\rot{MIG-sup} & 
\rot{JEMMIG} & 
\rot{Modularity Score} & 
\rot{DCIMIG} \\
$\theta \to [\textup{cos}\,\theta, \textup{sin}\,\theta]$ 
& 1.0 & 1.0 & 1.0 & 0.8 & 0.8 & 1.0 & 0.6 & 1.0 & 0.7 & 1.0 & 1.0 & 0.6 & 0.0 & 0.7 & 0.4 & 1.0 & 0.6 \\
$\theta \to [\theta, \theta]$ 
& 1.0 & 1.0 & 1.0 & 0.9 & 1.0 & 1.0 & 1.0 & 1.0 & 0.7 & 1.0 & 1.0 & 0.0 & 0.0 & 1.0 & 0.5 & 1.0 & 1.0 \\
$\theta \to [\theta, \theta, \theta, \theta]$
& 1.0 & 1.0 & 1.0 & 0.9 & 1.0 & 1.0 & 1.0 & 1.0 & 0.4 & 1.0 & 1.0 & 0.0 & 0.0 & 1.0 & 0.5 & 1.0 & 1.0 \\
\end{tabular}}
\label{table_angle}
\end{table*}

\subsection{Nonlinear relations}
\label{exp_nonlinear}
\begin{figure}[ht]
     \centering
     \begin{subfigure}[b]{0.35\textwidth}
         \centering
         \includegraphics[width=\textwidth]{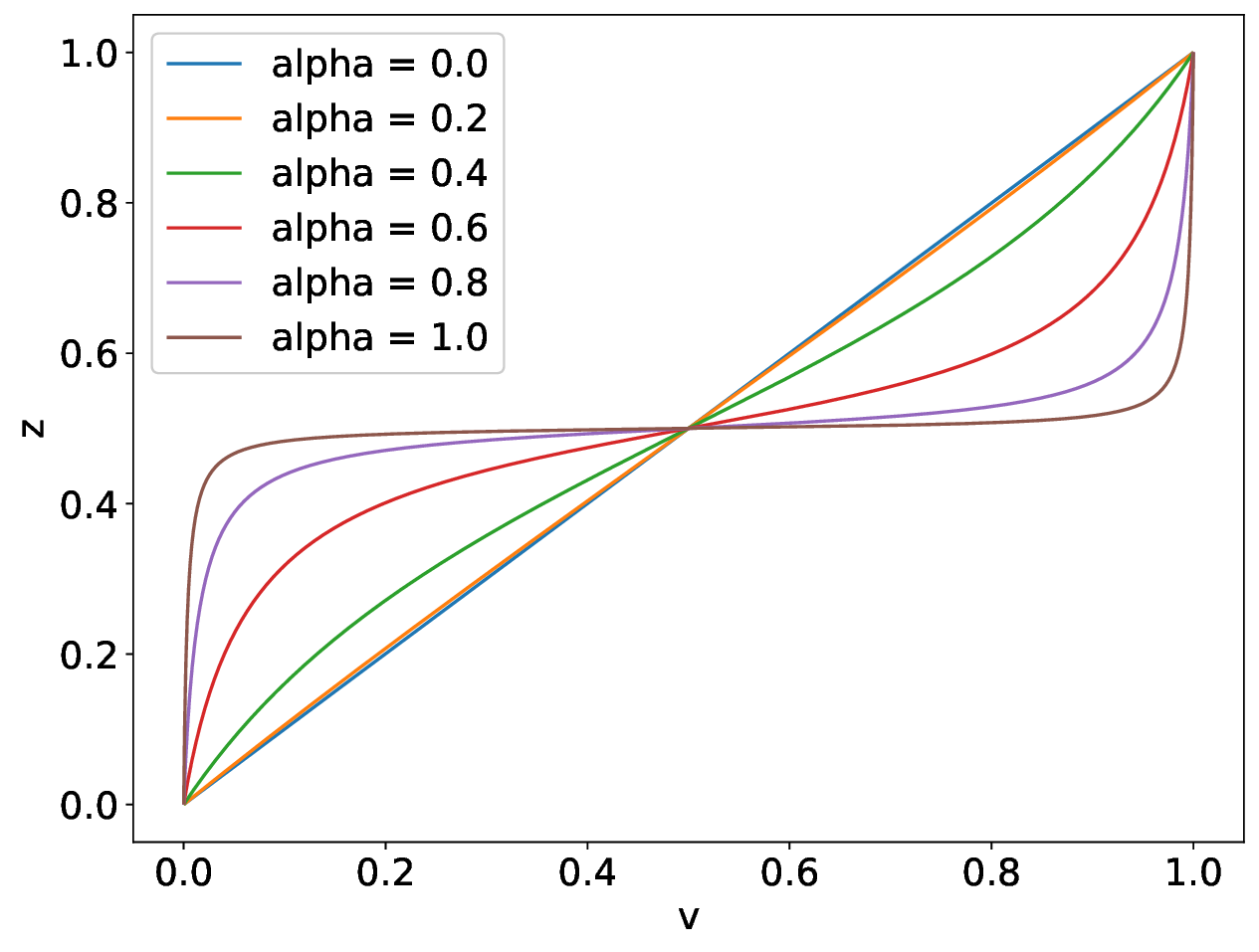}
         \caption{Shape of the factor-code relation as parameter $\alpha$ increases.}
         \label{fig_tan_example}
     \end{subfigure}
     \begin{subfigure}[b]{0.35\textwidth}
         \centering
         \includegraphics[width=\textwidth]{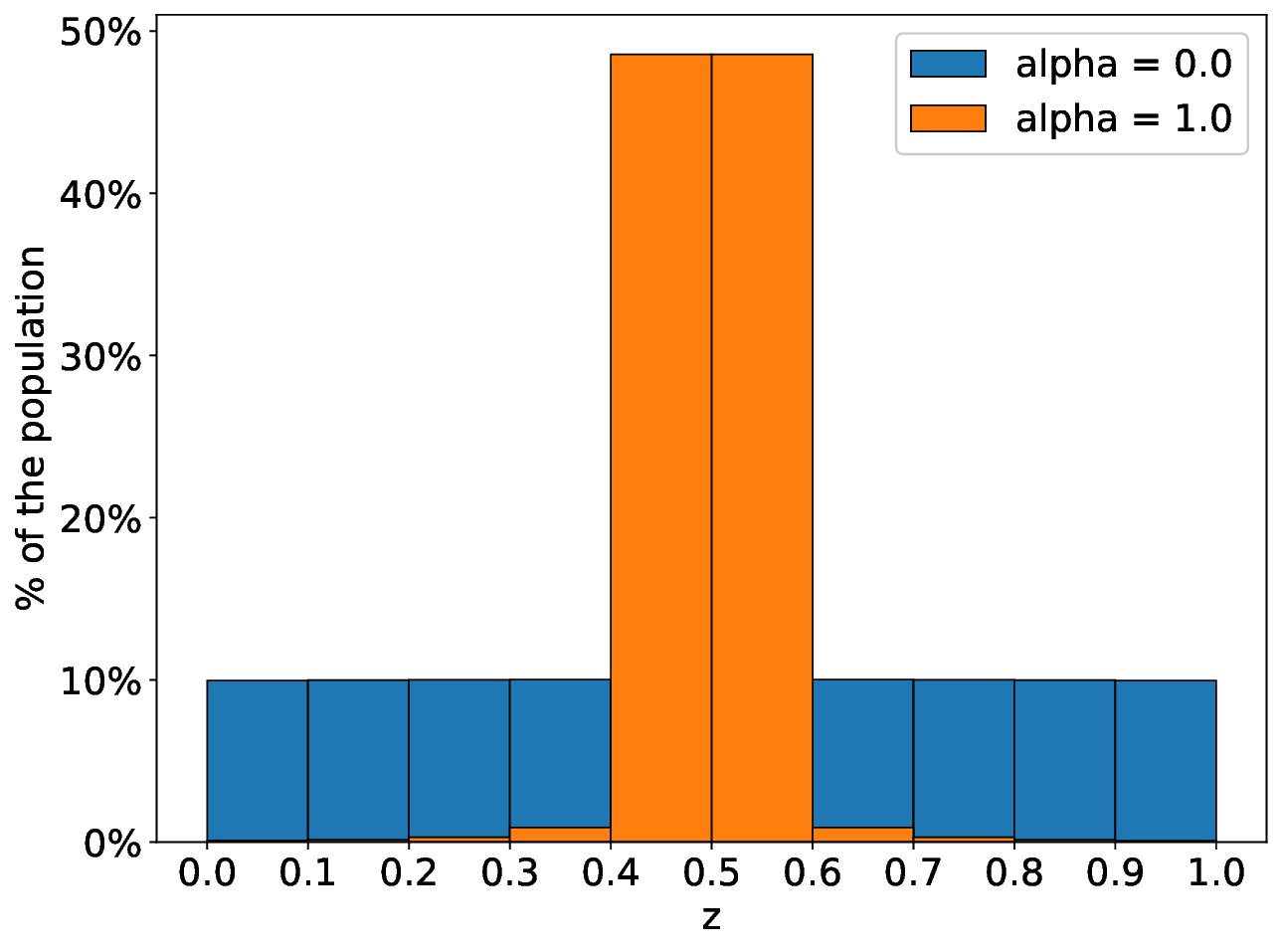}
         \caption{Effect of nonlinearity on discretization bins population.}
         \label{fig_tan_bins}
     \end{subfigure}
\caption{Shape of the parametric factor-code relation and its effect on discretization}
\end{figure}

\begin{figure*}[!ht]
\centering
\includegraphics[width=\textwidth]{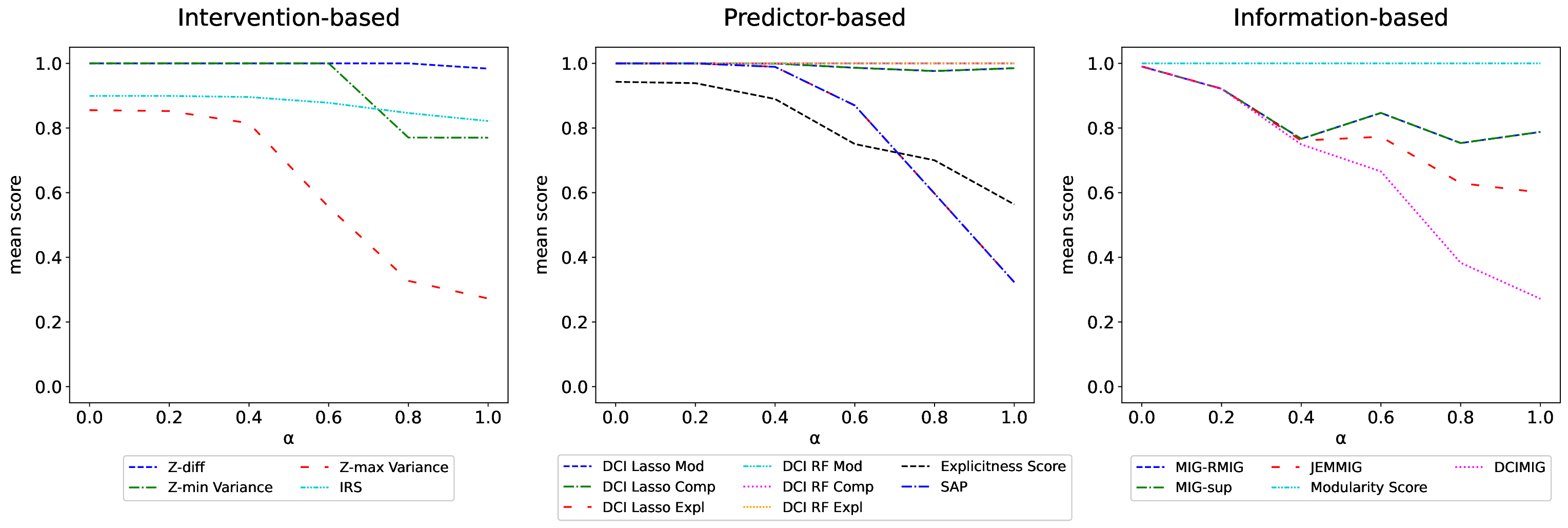}
\caption{Metric scores for perfectly disentangled representations with increasingly nonlinear factor-code relation ($\alpha$).}
\label{fig_result_non_linear}
\end{figure*}

Here we explore representations with nonlinear relations between factors and codes. The representation is kept perfectly compact and modular and should receive a perfect score from all metrics. The mapping function becomes increasingly nonlinear as $\alpha$ increases, but is always monotonic for $v \in [0,1]$:
\begin{equation}
    \textbf{\textup{z}} = f(\textbf{\textup{v}}) = 1000^{-\alpha+0.25} \, \textup{tan}(\omega (\textbf{\textup{v}}-0.5)) + 0.5
\end{equation}
where $\omega = 2\,\textup{arctan}(1000^{\alpha-0.25}/2)$. When $\alpha$ = 0, the relation is practically linear, and when $\alpha$ is 1 the relation takes the shape of a tangent function as shown in Figure \ref{fig_tan_example}. This relation is interesting because it highlights potential problems with using a linear regressor to compute scores, as well as potential problems inherent to discretization. Results are reported in Figure \ref{fig_result_non_linear}. 

As expected, predictor-based metrics using a linear regression to measure explicitness, DCI lasso and SAP, under-perform as the factor-code relation becomes less linear. The monotonic nature of the relation allows DCI lasso to accurately score modularity and compactness. Naturally, a more expressive predictor makes the metric robust to more complex relationships.

This experiment highlights potential problems with discretization which is at the center of information-based metrics, as well as intervention-based metrics and even some predictor-based metrics like the Explicitness score. Equal binning of the code space results in a larger amount of the population being assigned to the middle bins. Figure \ref{fig_tan_bins} shows the proportion of samples assigned to each discretization bins when $\alpha = 0$ and $\alpha = 1$. This uneven population distribution lowers the code space entropy and in turn affects MI computation. Similarly, it affects how subsets are created in intervention-based metrics. This explains why a large proportion of metrics fail to properly score the perfectly modular, compact and explicit representation.

\subsection{When Factors Partially Describe Data}
\label{exp_partial_desc}
This experiment simulates the case where metrics measure only a fraction of all the generative factors. This frequently happens in real-world scenarios because it is difficult to identify all generative factors in a data set. For instance, channel noise may corrupt data and get modeled in some dimensions of the code as in \cite{Hsu2019}. These non-measured factors still need to be encoded to preserve explicitness and because they can be useful for downstream tasks. 

From the metric point-of-view, code dimensions corresponding to non-measured factors are seen as noise or dead-codes \cite{Eastwood2018} which affects metric scoring. We generate perfect representations in the same way as in Section \ref{perfect_plus_noise} but without noise ($\alpha=0$). The relation between factors and codes becomes the identity $\textbf{\textup{z}} = f(\textbf{\textup{v}}) = \textbf{\textup{v}}$. Then, we apply the metrics to these perfect representations and vary the proportion of measured factors. 

\begin{figure*}[!ht]
\centering
\includegraphics[width=\textwidth]{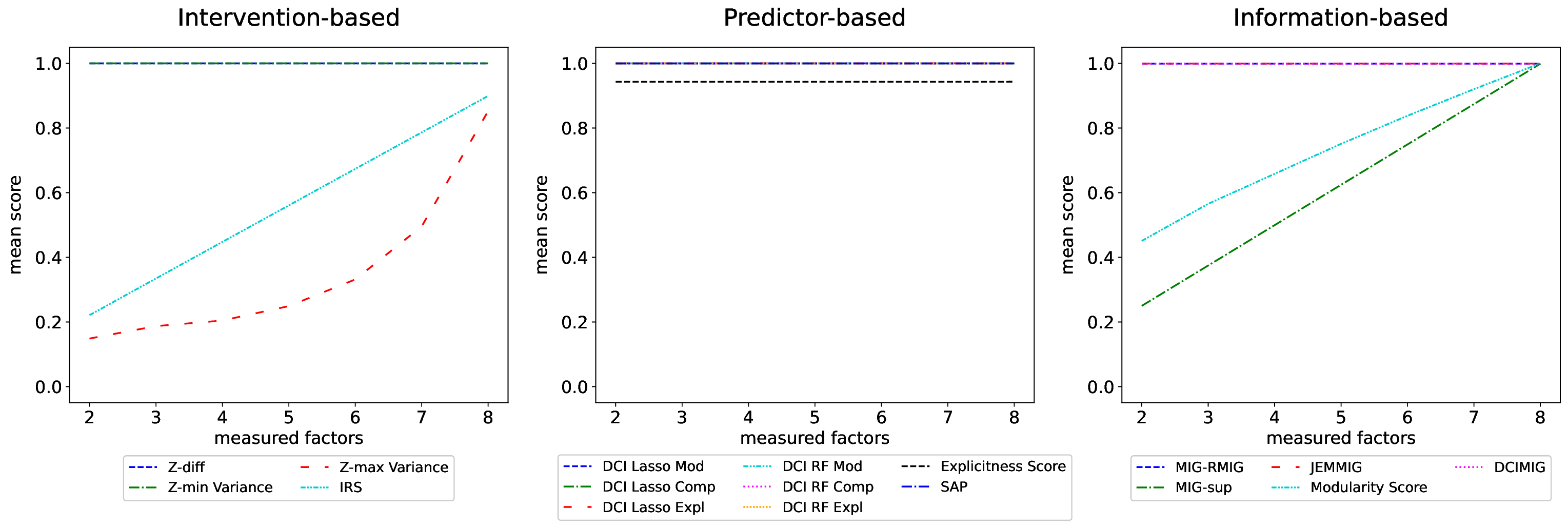}
\caption{Scores for perfectly disentangled representations. The abscissa indicates how many of the 8 factors are measured by the metrics.}
\label{fig_all_fractions}
\end{figure*}

When metrics measure all of the 8 generative factors captured by the perfect representation, their score should be maximal. Metrics should maintain that maximal value as the proportion of measured factors decreases because the representation does not change. Figure \ref{fig_all_fractions} shows how metric scores evolve as the proportion of measured factors decreases.

Most metrics are equipped to deal with non measured factors, except for Z-max Variance, IRS, MIG-sup and the Modularity score. This limits their relevance in contexts outside of academic toy problems. Successful methods that measure disentanglement from the code point-of-view implement a mechanism to discard \textit{dead-codes} \cite{Eastwood2018}. A dead-code is a code dimension that does not inform on any factor. The Modularity score does not provide such mechanisms. 
Also our implementation of MIG-sup does not account for dead-codes because it requires tuning a rejection threshold, which is not applicable in practice. Borrowing strategy to deal with dead-codes from other metrics would be beneficial as discussed in Section \ref{sect_migsup}. When sampling to create subsets, IRS and Z-max Variance implicitly assume that when all known factors are fixed, corresponding codes are also fixed. This assumption is violated when there are other sources of variation for the code than the known factors.


\subsection{Sample Efficiency}
\label{exp_sample_efficiency}
This section studies how many data points the metrics need to get a fair estimation of the representation score. The intuition is that an ideal metric should attribute the same score regardless of the number of samples observed for the same representation. While it does not make sense to expect a fair estimation of the true score without a minimal quantity of samples, this minimal quantity differs across metrics.

\begin{figure*}[!ht]
\centering
\includegraphics[width=\textwidth]{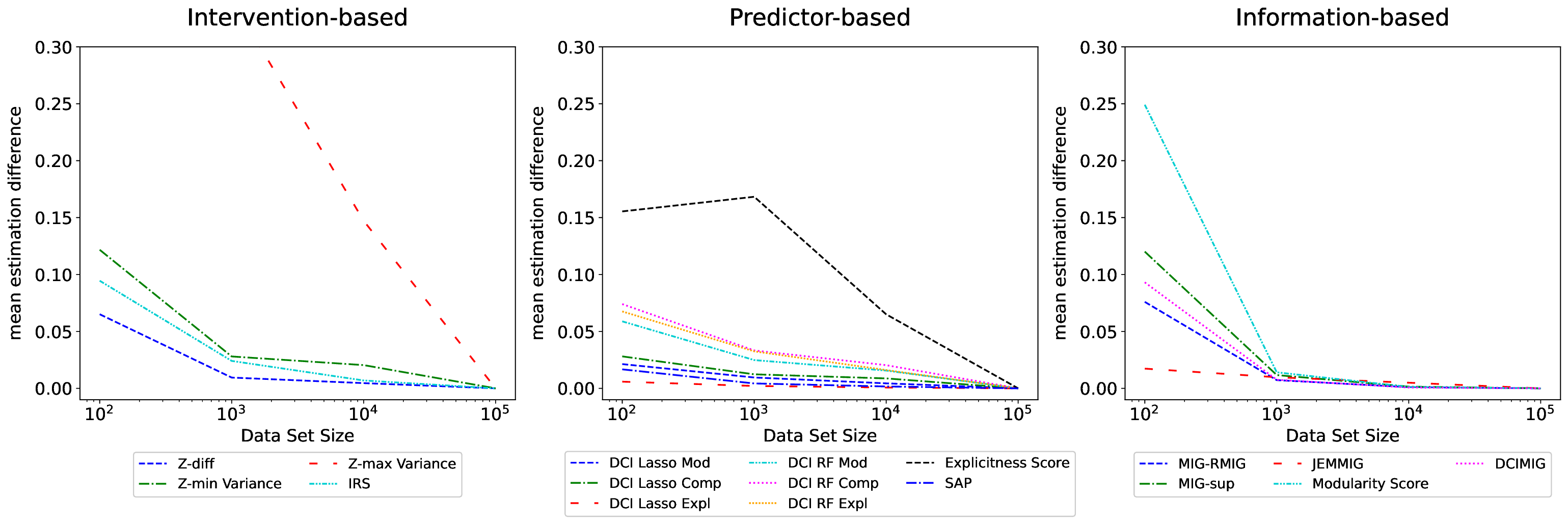}
\caption{Evolution of the score estimation quality with respect to the number of data points. The abscissa is the size of the sample subset used by the metrics to evaluate a representation. The ordinate is the mean of the score differences when compared with scores obtained with the full 100k data points for the same representation.}
\label{fig_sample_efficiency}
\end{figure*}

In the experiment, we create random representations: $\textbf{\textup{z}} = f(\textbf{\textup{v}}) = \textbf{\textup{v}}R$ where the projection matrix $R$ is filled by sampling in the uniform distribution and each element is replaced by 0 with a 0.75 probability. We create these random representations to avoid biases that could be created by perfect, or completely noisy representations. We generate 100k samples for the representation. Then, we apply each metric to the first 100, 1k and 10k samples and finally to the whole 100k sample set. We compute the absolute difference between each scores and the full 100k samples score. We repeat this process 100 times and average the differences. We report the results in Figure \ref{fig_sample_efficiency}.

All metrics provide a fair estimate of the disentanglement score 1k samples except for Z-max Variance and the Explicitness Score. The Z-max Variance metric requires a large amount of data because of the way it construct batches on which it measures code variance. In a batch all factors except one are fixed. This entails that the data set must contain several examples in which the majority of the code is the same, which is only possible with extremely large data sets. In our experiment there are 7 fixed code dimension each quantized to 10 bins. It is difficult to cover the $10^7$ code possibilities with only $10^5$ samples and obtain a reliable estimate. In the Explicitness Score the distribution of samples across its internal classes affect the shape of the multiple ROC curves leading to an unstable estimation without large data sets. 

Intervention-based metrics rely on sampling to create batches which necessitates large number of samples to provide a reliable estimate. Predictor-based metrics necessitate sufficient data to train a predictor that generalizes. The quantity of needed data points depends on the predictor model complexity. Simpler models such as linear regressions require less data than more complex models like random forests. This explains the gap observed between the DCI Lasso and DCI RF. On the other hand, Information-based metrics are generally more sample efficient because they are free from stochastic components. For all metrics, the minimal quantity of data needed to get a fair estimates depends on the number factor and their distribution in the data set, as well as the latent space dimensionality.

\section{Discussion}
\label{section_discussion}
This section summarizes our learnings from the experiment results and insights from relevant papers. After discussing relations between representation properties, we identify best practices for measuring disentanglement in real-world applications. Finally, we provide recommendations on how measurements should be reported.

\subsection{Relations Between Representation Properties}
\commentmarc{subsection intro} While disentanglement properties can be measured separately, they are implicitly linked together. This makes the analysis of disentanglement more difficult and might have motivated the holistic approach of some metrics.

\commentmarc{explicitness with other properties}
Evidently, some degree of explicitness is necessary to observe modularity or compactness otherwise it would not be relevant to compare factor-code relations. However, as shown in the experiment of Section \ref{perfect_plus_noise}, a high level of modularity and compactness can be observed, even when explicitness is minimal. In other words, \textbf{explicitness is a necessary condition for modularity and compactness, but the magnitude of these properties does not inform on the magnitude of the explicitness.}

\commentmarc{link between compactness and modularity}
\textbf{Modularity and compactness are linked together by the size of the code space.} When all factors are represented, if the code space is the same dimensionality as the number of factors, perfect modularity necessarily implies perfect compactness. This relation is not symmetric. Perfect compactness does not necessarily means perfect modularity in this situation. A code dimension could encode two factors even if each factor is encoded by only one dimension. This would however mean that there are one or more dead-codes. As the code space increases, under perfect modularity, imperfect compactness is possible. The ratio between the code space dimensionality and the number of factors determines how much compactness is allowed to deteriorate. As a general rule, the code space should be larger than the \textit{measured} factor space to allow for composite factors and non-measured factors. The fact that the code space size links compactness and modularity could explain correlations sometimes observed between metrics that focus on only one of these properties as in \cite{Locatello2019challenge}.

\commentmarc{Modularity vs. Compactness}
\textbf{Modularity is more important in practice than compactness.} This has already been stated in \cite{Ridgeway2016}. There are several reasons why researchers and practitioners should focus on modularity instead of compactness. The main reason is that measuring compactness is desirable only if one can identify \textit{atomic} (1D) factors, which is often difficult or impossible in real-world applications. As mentioned earlier, some basic concepts like angle or color are best represented in a 2D or 3D space. In addition, any composite factors (e.g., object type in images or speaker identity in speech segments) are more meaningfully represented in a multi-dimensional space, where each dimension represents an atomic factor. These atomic factors are sometimes concepts difficult to identify, describe and measure. For example, if one wants to disentangle speaker identity from speech segments. Many atomic factors define a voice print. Some are simpler to identify and measure like pitch and speech rate, but they do not paint the whole picture. The complete set of atomic factors for voice print remains elusive, even for speech experts. Nonetheless, they must be encoded to successfully perform a downstream task like speaker identification or conditioned speech synthesis. The same goes for illumination in a picture. In practice, one might want to isolate the effect of a light source. However, light sources have many attributes such as 3D position, direction, color, shape, size and intensity. All these factors have to be explicitly identified and quantified to measure compactness. For both these example applications, useful disentanglement is best measured through modularity. A high modularity score indicates that atomic factors of interest are contained in a defined subset of the code space.

\commentmarc{Alignement}
\textbf{The factor space and the code space need to be aligned to accurately measure disentanglement.} Perfect compactness and modularity entails complete disentanglement of generative factors. However, it is possible to learn a representation where factors are completely disentangled, and yet measure low compactness and modularity scores because of a misalignment between space axes. As a thought experiment, if we take a perfectly disentangled representation, where each factor corresponds to only one code, and rotate this representation space around any axis. The resulting rotated representation space maintains the independence between factors. However existing metrics will fail to capture perfect modularity or compactness because variations in one factor will cause variation on several dimensions of the code space and \textit{vice versa}. We believe a metric should be robust to this kind of misalignment and be able to evaluate representations by looking at them from the right "point-of-view". In practice, axis alignment can be enforced during learning through supervision, or indirectly encouraged as in VAEs \cite{Burgess2017, rolinek2019variational}, but cannot be guaranteed in most unsupervised learning settings \cite{Locatello2019challenge}.

\begin{table*} \centering

\caption{Summary of findings from experiments and analysis. For a metric to possess a desired characteristic (\cm), it has to be true in theory, as well as in practice. The robustness to noise characteristic does not apply to explicitness metrics.}
\scalebox{1.0}{
\begin{tabular}{lcccccccccccc}

\textbf{Metric} & 
\rot{Modularity} & 
\rot{Compactness} & 
\rot{Explicitness} & 
\rot{Calibrated} &
\rot{Robust to Noise} &
\rot{Robust to } \rot{Non-measured Factors}&
\rot{Nonlinear Relation} & 
\rot{Discretization-free} & 
\rot{Few Hyper-parameters} & 
\rot{Data Efficient}\\

\cmidrule[1pt]{1-11}
Z-diff \cite{Higgins2016}&
\xm & \xm & \xm & \cm & \cm & \cm & \xm & \xm & \xm & \cm\\

Z-min Variance \cite{Kim2018}&
\xm & \xm & \xm & \cm & \cm & \cm & \xm & \xm & \xm & \cm\\

Z-max Variance \cite{Kim2019}&
\xm & \xm & \xm & \xm & \xm & \xm & \xm & \xm & \xm & \xm\\

IRS \cite{Suter2019}&
\cm & \xm & \cm & \xm & n/a & \xm & \xm & \xm & \xm & \cm\\

\cmidrule{1-11}

DCI - Lasso \cite{Eastwood2018}&
\cm & \xm & \cm & \xm & \cm & \cm & \xm & \cm & \cm & \cm\\

DCI - Random Forest \cite{Eastwood2018}&
\cm & \cm & \cm & \cm & \cm & \cm & \cm & \cm & \xm & \cm\\

Explicitness Score \cite{Ridgeway2018}&
\xm & \xm & \cm & \xm & n/a & \cm & \xm & \xm & \cm & \xm\\

SAP \cite{Kumar2018}&
\xm & \cm & \cm & \cm & n/a & \cm & \xm & \cm & \cm & \cm\\

\cmidrule{1-11}

MIG-RMIG \cite{Chen2018, Do2020}&
\xm & \cm & \xm & \cm & \xm & \cm & \xm & \xm & \cm & \cm\\

MIG-sup \cite{Li2020}&
\cm & \xm & \xm & \cm & \xm & \xm & \xm & \xm & \cm & \cm\\

JEMMIG \cite{Do2020}&
\cm & \cm & \cm & \cm & \xm & \cm & \xm & \xm & \cm & \cm\\

Modularity Score\cite{Ridgeway2018}&
\xm & \xm & \xm & \xm & \cm & \xm & \xm & \xm & \cm & \cm\\

DCIMIG \cite{Sepliarskaia2020}&
\cm & \xm & \cm & \cm & \xm & \cm & \xm & \xm & \cm & \cm\\

\cmidrule[1pt]{1-11}
\end{tabular}}

\label{Table:matrix}
\end{table*}

\subsection{Practical Considerations for Choosing a Metric}
\label{section_discussion_pratical}
In this section, we extract conclusions from our analysis and experimental results. We provide guidance for choosing an appropriate metric for real-world applications, and we highlight practical considerations when measuring disentanglement.

Table \ref{Table:matrix} compiles our experimentation results and analysis. For a metric to possess a characteristic (\cm), it has to be true by design and not disproven experimentally. For instance, DCI with lasso regressor is marked with (\xm) because it has a failure mode when measuring compactness as shown in Section \ref{exp_modular_not_compact}, even if in theory it can measure the property. Same goes for metrics necessitating discretization when dealing with nonlinear relations.

\textbf{Metrics that do not account for non-measured factors should be avoided in real-world scenarios.} As discussed in Section \ref{factor_ind}, identifying factors in practice is challenging. Identifying all factors is even more difficult. Moreover, when identified, factors must be measured which is sometimes impossible. This means that for most applications there will exist unidentified factors explaining the data, which will cause some metrics to underestimate modularity as shown in Section \ref{exp_partial_desc}. 

\textbf{Using discretization is not trivial and has an impact on score.} As we saw in Section \ref{exp_nonlinear}, discretization of the code and the factor space has considerable impact on the ability of metrics to deal with nonlinear relations. 

The granularity of the discretization has an impact on the estimated MI, which is the centerpiece of information-based metrics. Intuitively, MI informs on how easy it is to predict a variable $A$ knowing $B$. Suppose $A$ is a random variable and $B = A + \sigma$ where $\sigma$ is random noise. On one extreme, if both variables are discretized in 1 bin, then the MI is maximum. On the other end of the spectrum, $A$ and $B$ are discretized in a large number of narrow bins. If the number of samples is limited, it is unlikely that $B$ will help predict the exact bin of $A$. In that case, MI will appear to be low even if there exists a strong relation between $A$ and $B$. This being said, when representation distributions are simple, MI can be analytically computed and these considerations can be avoided. 

In intervention-based metrics, the discretization granularity determines the degree of similarity/dissimilarity of examples grouped in the same subset. A too coarse discretization creates heterogeneous groups that are considered homogeneous, which biases results. A too fine discretization makes it impossible to create large enough subsets of data points with the same fixed value. To our knowledge, no procedure has been proposed yet to strike the right balance between coarse and fine discretization for any type of metric.

\textbf{DCI implemented with random forest is the best all around metric.} Measuring disentanglement properties separately allows for accurate scoring. Because random forest is an expressive model, it can discover nonlinear relationships and does not suffer from problems related to discretization. Moreover, random forests can be used as classifiers and regressors which makes them appropriate for applications mixing continuous and categorical factors. DCI implements a weighting scheme that accounts for dead-codes in problems where not all factors can be identified. However, there are three disadvantages to DCI. First, modeling relations with random forests requires a bit of expertise to set the hyper-parameters and determine a relevant criterion for code dimension importance. \textbf{The hyper-parameters must be tuned using an appropriate cross-validation procedure, to ensure proper regularization of the model.} Otherwise it will overfit, which results in an overestimation of explicitness as well as an underestimation of modularity and compactness. This cross-validation procedure is time consuming which is the second main disadvantage of the method. In fact, DCI with random forest is by far the most computationally expensive of all metrics implemented in this paper. Finally, training reliable RF models requires appreciable quantity of data points when compared to some other metrics.

In their current state, metrics in the intervention-based family should be used with great caution. They require large quantities of data to create subsets with fixed values. This prohibits their application in problems with limited quantities of data with labeled factors. They are subject to vulnerabilities associated with discretization. Moreover, they are prone to failure modes, which limits their reliability. Finally, unlike most metrics from other families, they do not produce a factor-code relation matrix, which makes their results difficult to interpret and less helpful when debugging.

Information-based metrics are in theory flexible and elegant. They can measure factor-code relations of any shape, continuous or categorical, with a minimal amount of hyper-parameter tuning and few data points. However, the aforementioned challenges with discretization limit their universality and makes them vulnerable to noise. Also metrics based on information \textit{gaps} like MIG, only consider the difference between the two best candidates. This limits their expressiveness. For instance in the experiment of Section \ref{exp_modular_not_compact}, MIG attributes the same compactness score (0.0) to representations where a factor corresponds to two and four code dimensions. We believe that if these limitations were addressed, information-based metrics would be more interesting solutions.

\subsection{Reporting Results}

\textbf{Disentanglement properties should be measured separately.} We share this opinion with \cite{Eastwood2018} and \cite{Ridgeway2018}. In our experiments, we showed we could vary properties independently and get the same overall score in very different situations. Metrics measuring all at once make the analysis and comparison of algorithms imprecise. This is particularly true in cases where a parameter balances reconstruction error and factor separation (e.g. $\beta-$VAE \cite{Higgins2016}). Using a single metric to measure both explicitness and modularity makes it impossible to determine the contribution of each property to the score. 

\textbf{Disentanglement should be measured for each factor independently.} While global scores give a quick impression on disentanglement quality, they do not paint the whole picture and can be deceiving. It is impossible to tell from a single number if a model performs generally well except on a few problematic factors, or equally badly on all of them. The first case might indicate a problem with the data or the choice of factors, while in the second it indicates poor performance of the representation model.

\textbf{Metrics should be run several times on the same representation. One should report average scores alongside standard deviation.} Some metrics implement stochastic components. For instance, intervention-based metrics sample subsets on which they rest their analysis. Predictor-based metrics create validation sets to perform hyper-parameter tuning. Moreover, in applications with large data sets, representations are evaluated on a subset of samples for efficiency. This sampling process adds to the stochasticity of the evaluation even for stable metrics. Performing several measurement runs allows performing statistical significance tests on results to ascertain conclusions from experiments, which should be standard practice when comparing different solutions.

\textbf{A minimal sample set size is required to get an accurate estimation of the disentanglement score of a learned representation.} As explained in Section \ref{exp_sample_efficiency} this minimal quantity depends for each metric, factor distribution and code space dimensionality. For instance in our experiment metrics such as DCI RF and Z-min Variance provide an estimate of the true score that most likely differs by $\pm 0.03$ from the representation \textit{true} score, even with $N=10 000$, while Information-based metrics fare better than their counterparts when fewer samples are available. This score estimation error should be taken into account when comparing representations and calls for caution when drawing conclusions. 

\section{Conclusion}

In this work we studied how to quantify disentanglement in representations. We conducted an extensive review of supervised disentanglement metrics. We analyzed and compared them experimentally with real-world applications in mind. We reviewed definitions of disentanglement and proposed a new taxonomy organizing the metrics into three families: intervention-based, predictor-based and information-based. 

We highlighted the lack of correlation between the different metric scores, and exposed their differences in a series of fully controlled experiments on the robustness to noise, modularity, compactness, hidden factors, calibration and nonlinear relationships. Our experiments revealed different limitations for each metric. We showed how discretization hinders reliability under limited amount of data, noise and nonlinear factor-code relations. We found that predictor-based metrics, when parameterized with caution, were the best performing family of solutions. We discussed the importance of modularity over compactness for practical applications. We concluded, perhaps unsurprisingly, that each disentanglement property should be measured separately for better interpretability.

While we shed some light on the inner working assumption of supervised metrics, several open questions remain. We think that some of the limits exposed in the study can be solved, and thus some metrics, notably from the information-based family, could prove to be stronger solutions than they are now. Also, supervised metrics necessitate factors to be identified and measured which is not always possible when dealing with real-world data. This is why research efforts are now increasingly focused on measuring disentanglement without ground truth factors. This study intentionally left out unsupervised metrics, which is open for future work.

\bibliographystyle{ieeetr}  
\bibliography{main}  


\end{document}